%% file: paper_arxiv.tex
\documentclass{article}

\usepackage{amsmath,amsthm,amssymb,amsfonts}

\usepackage{hyperref}
\hypersetup{
    pdfnewwindow=true,     
    colorlinks=true,     
    linkcolor=blue,     
    citecolor=blue,     
    filecolor=magenta,     
    urlcolor=cyan     
}

\usepackage[capitalize]{cleveref}
\usepackage{enumitem}
\usepackage{cancel}
\usepackage{mathtools}
\usepackage{bbm,dsfont}
\usepackage{graphicx}
\graphicspath{ {./plots/} }

\usepackage{natbib}
\newtheorem{theorem}{Theorem}

\input{commands-omar}

\newcommand{\bc}{\begin{center}}
\newcommand{\ec}{\end{center}}

\usepackage[textsize=tiny,
]{todonotes}
    \usepackage[preprint]{neurips_2019}

\title{PAC-Bayes with Backprop}

\author{%
  Omar Rivasplata \\
  DeepMind \\
  \texttt{rivasplata@google.com} \\
  \And
  Vikram Tankasali \\
  DeepMind \\
  \texttt{tvikram@google.com} \\
  \And
  Csaba Szepesvari \\
  DeepMind \\
  \texttt{szepi@google.com} \\
}

\begin{document}  

\iftrue
\maketitle
\fi

\begin{abstract}
We explore the family of methods `PAC-Bayes with Backprop' (PBB)
to train probabilistic neural networks by minimizing PAC-Bayes bounds. 
We present two training objectives, one derived from a previously known PAC-Bayes bound, and a second one derived from a novel PAC-Bayes bound. 
Both training objectives are evaluated on MNIST and on various UCI data sets. 
Our experiments show two striking observations: we obtain competitive test set error estimates ($\sim 1.4\%$ on MNIST) and at the same time we compute non-vacuous bounds with much tighter values ($\sim 2.3\%$ on MNIST) than previous results.
These observations suggest that neural nets trained by PBB may lead to self-bounding learning, where the available data can be used to simultaneously learn a predictor and certify its risk, with no need to follow a data-splitting protocol.
\end{abstract}

\setlength{\marginparwidth}{10ex}

\section{Introduction}
\label{s:intro}

In a probabilistic neural network the weights are random outcomes of a probability distribution, rather than having fixed deterministic values. For instance this may be realized by injecting noise into the components of a fixed architecture. The justification is that a probabilistic neural network may lead to better predictions, overcoming the limited generalization ability of a single weight vector (\cite{neal1993bayesian}). 
Other arguments in favour of weight randomization might be enforcing or studying robustness properties (\cite{NoyCrammer2014robust}, \cite{blundell2015weight}, \cite{zhang2019all}), and making available better tools for the design and analysis of learning algorithms. 

In this paper we take on probabilistic neural networks from a PAC-Bayes point of view.
We focus on the family of methods `PAC-Bayes with Backprop' (PBB) which derives training objectives based on PAC-Bayes upper bounds on the risk. 
%
%
The notable contributions of \cite{dziugaite2017computing,dziugaite2018data} showed that exploring this approach is totally worth the while.
Computational considerations aside, a clear advantage of PBB methods is being an instance of learning with guarantees: When training neural nets by PBB methods the output of the optimization process is a (randomized) predictor and simultaneously a risk certificate that guarantees its performance on unseen examples.
Naturally, learning with guarantees \emph{per se} will not impress until the values of the reported risk certificates match or closely follow the error estimates calculated on a test set, so that the former can be considered to be informative of the performance on unseen examples.

The results presented in this paper indicate a positive step
towards learning with guarantees. 
We report experimental results on MNIST with two training objectives.
The first training objective follows the PAC-Bayes-$\lambda$ bound of \cite{thiemann-etal2017}.
The second training objective comes from what we call the PAC-Bayes-quadratic bound, whose derivation involves solving a quadratic inequality, hence the name.
Our conclusions from these experiments are (1) that PBB methods achieve competitive (e.g. comparable to \cite{blundell2015weight}'s) test set error estimates ($\sim 0.014$ on MNIST), while (2) the minimal value of the upper bound that we achieve ($\sim 0.023$ on MNIST) is a significant improvement over previous (e.g. \cite{dziugaite2017computing,dziugaite2018data}'s) results, i.e. we further close the gap between the risk bound certificate and the test set risk estimate.

\paragraph{Our contributions:}
\begin{enumerate}[leftmargin=*,topsep=2pt,itemsep=2pt]
    \item Proposing two new training objectives for NNs: one derived from the PAC-Bayes-lambda bound, and one derived from the novel PAC-Bayes-quadratic bound.
    \item Connecting \cite{blundell2015weight}'s training method, which achieved competitive test set performance, to PAC-Bayes methods, which additionally produce a risk certificate and overcome some limitations of the former method, while also achieving competitive test set error estimates.
    \item Demonstrating via experimental results that the family of methods "PAC-Bayes with Backprop" might be able to achieve self-bounding learning: obtaining competitive test set error estimates and simultaneously computing non-vacuous bounds with much tighter values than previous works.
\end{enumerate}
We would also like to highlight the elegant simplicity of the training methods we present here: our results are achieved with the classical fixed `data-free' priors and classical SGD optimization (in contrast, \cite{dziugaite2018data} used a special kind of data-dependent priors, and optimization via SGLD). Besides, our methods do not involve tampering with the training objective (in contrast, \cite{blundell2015weight} used a ``KL killing trick'' by inserting a tunable parameter in their objective). This does not imply that extra resources do not have a valuable effect, but rather makes the point that it is indeed worthwhile studying simple methods, not just to understand their scope but also to more accurately assess the real value added by the extra resources.


Another set of experiments was run on five UCI datasets with the same two training objectives as our experiments on MNIST, again showing tightness of the gap between risk certificate and test set risk estimate. However, some initial experiments (not reported here) on CIFAR-10 gave risk certificates of three times the corresponding test set error estimate, which indicates that more needs to be done before `self-bounding learning' can really be claimed.

The rest of this paper is organized as follows. In \cref{s:prelim} we briefly recall some notions of the supervised learning framework, mainly to set the notation used later. Then in \cref{s:PAC-Bayes} we discuss the PAC-Bayes framework and two PAC-Bayes bounds. \cref{s:PBB} presents two training objectives derived from the PAC-Bayes bounds. \cref{s:BBB} discusses \cite{blundell2015weight}'s objective. In \cref{s:exp} we presents our experimental results for both methods. We conclude and discuss future research directions in \cref{s:conclude}. 
\cref{s:L_vs_G} displays two KL divergence formulas for convenience. 
\cref{s:exp_UCI} presents the results of another set of experiments on some UCI data sets.

\section{Generalization through risk upper bounds}
\label{s:prelim}

In the context of supervised learning, 
an algorithm that trains a neural network receives a finite list of training examples and produces a weight vector $w \in \mathcal{W} \subset \R^p$ which will be used to predict the label of unseen examples.
The ultimate goal is for the algorithm to find a weight that generalizes well, meaning that the decisions arrived at by using the learned $w$ should give rise to a small loss on unseen examples.%
\footnote{In statistical learning theory there is a precise meaning of when does a method generalize \citep{SSBD09}. We use the word generalization in a slightly broader sense here. 
} 
Turning this into precise statements requires some formalizations, briefly discussed next. The experienced readers should feel free to skip the next couple of paragraphs which aim to recall these formalizations as well as setting the notation for the rest of the paper.

The training algorithm receives a size-$n$ list of labelled examples $Z_{1:n} = (Z_1,\ldots,Z_n)$, where the examples $Z_i$ are randomly drawn from a space $\mathcal{Z}$ according to an unknown underlying probability distribution%
\footnote{$M_1(\mathcal{Z})$ denotes the set of all probability measures over $\mathcal{Z}$.}
$P \in M_1(\mathcal{Z})$. 
The form of the example space in supervised learning is $\mathcal{Z} = \mathcal{X}\times\mathcal{Y}$ with $\mathcal{X}\subset\R^d$ and $\mathcal{Y}\subset\R$, each example being a pair $Z_i = (X_i,Y_i)$ consisting of an input $X_i$ and its corresponding label $Y_i$. 
A space $\mathcal{W}$ encompasses all possible weight vectors, and
it is understood that each possible weight vector $w \in \mathcal{W}$ maps to a function $h_w: \mathcal{X}\to\mathcal{Y}$ that will assign a label $h_{w}(X) \in \mathcal{Y}$ to each new input  $X \in\mathcal{X}$. 
While statistical inference is largely concerned with learning properties of the unknown data-generating distribution, the main focus of machine learning is on the quality of predictors, measured by the expected loss on unseen examples, also called the risk:
\begin{align}
    L(w) = \EE[\ell(w,Z)] = \int_{\mathcal{Z}} \ell(w,z) P(dz)    \,.
\end{align}
Here $\ell : \mathcal{W}\times\mathcal{Z}\to[0,\infty)$ is a fixed loss function. 
With these components, regression is defined as the problem when $\cY = \R$ and the loss function is the squared loss $\ell(w,z) = (y-h_w(x))^2$,
while binary classification is defined with $\cY = \{0,1\}$ (or $\cY = \{-1,+1\}$) and setting the loss to the zero-one loss:  $\ell(w,z) = \mathbb{I}[y \neq h_w(x)]$.

The goal of learning is to find a weight vector with small risk $L(w)$.
Since the data-generating distribution $P$ is unknown, $L(w)$ is an unobservable objective. 
Replacing the expected loss with the average loss on the data gives rise 
to an observable objective called the \emph{empirical risk} functional:
\begin{align}
    \hat{L}_n(w)
    = \hat{L}(w,Z_{1:n}) 
    = \frac{1}{n}\sum_{i=1}^{n} \ell(w,Z_i)    \,.
\end{align}
In practice, the minimization of $\hat{L}$ is often done with some version of gradient descent. Since the zero-one loss gives rise to a piecewise constant loss function, in classification it is common to replace it with a smooth(er) loss, such as the cross-entropy loss, while changing the range of $h_w$ to $[0,1]$.



Under certain conditions, minimizing the empirical risk leads to a weight that is guaranteed to have a small risk. 
Examples of such conditions are when the set of functions $\{h_w \,:\, w\in \R^p\}$ representable has a small capacity relative to the sample size or the map that produces the weights given the data is stable, or
the same map is implicitly constructed by an algorithm such as stochastic gradient descent (see e.g. \cite{bottou2012sgd_tricks} and references).
However, oftentimes minimizing the empirical risk can lead to a situation when the risk of the learned weight is larger than desired -- a case of overfitting.
To prevent overfitting, various methods are commonly used. These include complexity regularization, early stopping, injecting noise in various places into the learning process, etc (e.g. \cite{srivastava2014dropout}, \cite{wan2013regularization}, \cite{caruana2001overfitting}, \cite{hinton1993keeping}).

An alternative to these is to minimize a surrogate objective which is guaranteed to give an upper bound on the risk.
As long as the upper bound is tight and the optimization gives rise to a small value for the surrogate objective, the user can be sure that the risk will also be small: In this sense, overfitting is automatically prevented, while we also automatically get a self-bounded method in the sense of  \citet{Freund1998} (see also \cite{langford2003microchoice}). 
In this paper we follow this last approach, with two specific training objectives derived from corresponding PAC-Bayes bounds, which we introduce in the next section.


\section{Two PAC-Bayes bounds}
\label{s:PAC-Bayes}

Probabilistic neural networks are realized as probability distributions over the weight space. While a classical neural network learns a data-dependent weight vector $\hat{w}$, a probabilistic neural network learns a data-dependent distribution over weights, say $Q \in M_1(\mathcal{W})$, 
and makes randomized predictions according to this distribution. To make a prediction on a fresh example $X$ the randomized predictor draws a weight vector $W$ at random according to $Q$ and applies $h_W$ to $X$: 
Each new prediction requires a fresh draw.
In this case the performance measures are lifted by averaging with respect to the randomizing distribution. The average empirical loss is
$Q[\hat{L}_n] = \int_{\mathcal{W}} \hat{L}_n(w) Q(dw)$
and the average population loss is
$Q[L] = \int_{\mathcal{W}} L(w) Q(dw)$.

Given two probability distributions $Q, Q^{\prime} \in M_1(\mathcal{W})$, the Kullback-Leibler (KL) divergence from $Q$ and $Q^{\prime}$, also known as relative entropy, is defined as follows:
\begin{align*}
\vspace{-4mm}
\KL(Q \Vert Q^{\prime}) = \int_{\mathcal{W}} \log\Bigl( \frac{dQ}{dQ^{\prime}} \Bigr) \,dQ\,.
\vspace{-4mm}
\end{align*} 
Here $dQ/dQ^{\prime}$ denotes the Radon-Nikodym derivative. Clearly the KL gives a finite value when $Q$ is absolutely continuous with respect to $Q^{\prime}$.
For Bernoulli distributions with parameters $q$~and~$q^\prime$ we will write
$\kl(q \Vert q^\prime) = q \log(\frac{q}{q^\prime}) + (1-q)\log(\frac{1-q}{1-q^\prime})$, also called the binary KL divergence.

The PAC-Bayes-kl theorem (\cite{LangfordSeeger2001}, \cite{seeger2002}, \cite{maurer2004note}) concludes that, with high probability, the binary KL from $Q[\hat{L}_n]$ and $Q[L]$ is upper bounded as follows:
\begin{align*}
    \kl(Q[\hat{L}_n] \Vert Q[L]) \leq
    \frac{\KL(Q \Vert Q^{0})+\log(\frac{ 2\sqrt{n} }{\delta})}{n}\,.
\end{align*}
Inversion of the binary KL divergence 
based on the inequality $\kl(\hat{p} \Vert p) \geq (p - \hat{p})^2/(2p)$ valid for $\hat{p}<p$ (see e.g. \cite[Lemma 8.4]{boucheron2013concentration}) then gives:
\begin{align}
    Q[L] - Q[\hat{L}_n] \leq 
    \sqrt{ 
    2Q[L]\frac{\KL(Q \Vert Q^{0})+\log(\frac{2\sqrt{n}}{\delta})}{n}
    }\,.
\label{eq:invert}
\end{align}

On the one hand,
solving the quadratic inequality \eqref{eq:invert} for $\sqrt{Q[L]}$ leads to the following bound:

\begin{theorem}
\label{pb-quad}
For any $n$,
for any $P \in M_1(\mathcal{X})$,
for any $Q^{0} \in M_1(\mathcal{W})$,
for any loss function with range $[0,1]$,
for any $\delta \in (0,1)$,
with probability $\geq 1-\delta$ over size-$n$ i.i.d. samples, simultaneously for all $Q \in M_1(\mathcal{W})$ we have
\begin{align}
    Q[L] \leq \left(
    \sqrt{ 
    Q[\hat{L}_n] + \frac{\KL(Q \Vert Q^{0}) + \log(\frac{2\sqrt{n}}{\delta})}{2n} 
    }
    +
    \sqrt{ 
    \frac{\KL(Q \Vert Q^{0}) + \log(\frac{2\sqrt{n}}{\delta})}{2n} 
    } \right)^2\,.
\label{eq:quad-bound}
\end{align}
\end{theorem}

On the other hand, 
using \eqref{eq:invert} combined with the inequality $\sqrt{ab} \leq \tfrac{1}{2}(\lambda a + \frac{b}{\lambda})$ valid for all $\lambda > 0$, plus some derivations, leads to the PAC-Bayes-$\lambda$ bound of \cite{thiemann-etal2017}: 

\begin{theorem}
\label{pb-lambda}
For any $n$,
for any $P \in M_1(\mathcal{X})$,
for any $Q^{0} \in M_1(\mathcal{W})$,
for any loss function with range $[0,1]$,
for any $\delta \in (0,1)$,
with probability $\geq 1-\delta$ over size-$n$ i.i.d. samples, simultaneously for all $Q \in M_1(\mathcal{W})$ and $\lambda\in(0,2)$ we have
\begin{align}
Q[L]
\le 
\frac{Q[\hat{L}_n]}{1-\lambda/2}+ \frac{\KL(Q \Vert Q^{0})+\log(2\sqrt{n}/\delta)}{n\lambda(1-\lambda/2)}\,.
\label{eq:lambda-bound}
\end{align}
\end{theorem}

Notice that the conclusion of both theorems is an upper bound on $Q[L]$ that holds  simultaneously for all distributions $Q$ over weights, with high probability (over samples). The relevant case, of course, is when $Q$ is a data-dependent distribution, i.e. $Q$ learned from training data. On the other hand, in this context $Q^{0}$ is a fixed `data-free' distribution, external to the training process.

Below in \cref{s:PBB} we discuss two training objectives derived from these two bounds. Notice that there are many other PC-Bayes bounds available in the literature (references), which readily lead to corresponding training objectives.

\section{Two PAC-Bayes with Backprop (PBB) objectives}
\label{s:PBB}

The essential idea of ``PAC-Bayes with Backprop'' (PBB) is
to train a probabilistic neural network by minimizing an upper bound on the risk, specifically, a PAC-Bayes bound.
Here we present two training objectives, derived from \cref{eq:lambda-bound} and \cref{eq:quad-bound} respectively. 

Notice that the PAC-Bayes bounds (\cref{pb-lambda} and \cref{pb-quad}) require bounded losses, while neural network classifiers are trained to minimize the cross-entropy loss, which is unbounded. Hence in the experiments (below) we enforced an upper bound 
on the cross-entropy loss by lower-bounding the network probabilities by a value $p_\mathrm{min}>0$, so that the probabilities passed to the cross-entropy loss are $\max\{ p_\mathrm{min}, p_\mathrm{net} \}$, where $p_\mathrm{net}$ are the probabilities calculated by the network.
Using a surrogate loss, like the cross-entropy loss, is beneficial for making the error surface better behaved (continuous and piecewise differentiable), while it introduces a mismatch between the actual target (i.e. the misclassification loss) and the optimization objective.

The optimization objective derived from \cref{pb-quad} is:
\begin{align}
    f_{\mathrm{quad}}(Q) = 
    \left(
    \sqrt{ 
    Q[\hat{L}_n] + \frac{\KL(Q \Vert Q^{0}) + \log(\frac{2\sqrt{n}}{\delta})}{2n} 
    }
    +
    \sqrt{ 
    \frac{\KL(Q \Vert Q^{0}) + \log(\frac{2\sqrt{n}}{\delta})}{2n} 
    } \right)^2\,.
\label{eq:obj-quad}
\end{align}
The optimization objective derived from \cref{pb-lambda} is:
\begin{align}
    f_{\mathrm{lamb}}(Q) =
    \frac{Q[\hat{L}_n]}{1-\lambda/2}+ \frac{\KL(Q \Vert Q^{0})+\log(2\sqrt{n}/\delta)}{n\lambda(1-\lambda/2)}\,.
\label{eq:obj-lamb}
\end{align}

Optimization of \cref{eq:obj-lamb} is by alternating minimization with respect to $\lambda$ and $Q$, similar to the procedure \citet{thiemann-etal2017} used in their experiments with SVMs.
By choosing $Q$ appropriately, we use the pathwise gradient estimator as done by \cite{blundell2015weight}.

For the sake of clarification, we are not claiming to be the first to train a probabilistic neural network by minimizing a PAC-Bayes bound. 
However, as will be demonstrated below, our experiments based on the two training objectives above lead to (1) test set performance comparable to that of \cite{blundell2015weight}, while (2) computing non-vacuous bounds with tighter values than those obtained by \cite{dziugaite2017computing,dziugaite2018data}.
We do claim that our contributions are significant.

\section{The Bayes by Backprop (BBB) objective}
\label{s:BBB}

The `Bayes by backprop' 
of \cite{blundell2015weight} is inspired by a variational Bayes argument, which, in our notation, leads to the objective
\begin{align}
f(Q) = Q[\hat{L}_n] + \eta\, \frac{ \KL( Q \Vert Q^0 ) }{n}\,,
\label{eq:bbbobjective}
\end{align}
where 
 $Q^0$ plays the role of a prior distribution over the weights and
$\eta>0$, a hyperparameter, is introduced in a heuristic manner to make the method more flexible.
Note in particular that the variational Bayes argument gives $\eta=1$. 
When $\eta$ is treated as a tuning parameter, the method can be interpreted as searching in ``KL balls'' 
centered at $Q^0$ of various radii. Thus, the KL term then plays the role of penalizing the complexity of the model space searched.
\cite{blundell2015weight} propose to optimize this objective (for a fixed $\eta$) 
using stochastic gradient descent (SGD), which randomizes over both mini-batches (randomly selected subsamples of the training examples, \cite{bottou2012sgd_tricks}) and also over the weights and uses the so-called pathwise gradient estimate (\cite{price1958useful,jankowiak2018pathwise}).
This latter assumes that $Q = Q_\theta$ for $\theta\in \R^k$ with some $k>0$ is such that sampling from it can be accomplished by a smooth $\theta$-dependent transformation of a random variable sampled from a \emph{fixed} distribution $P_0$ while the density of $Q_\theta$ (with respect to some fixed reference measure) is also available in closed form. They propose to choose $P_0$ to be $p$-wise independent distribution and the transformation to act affine linearly component by component and argue that the resulting procedure has a computational cost similar to backpropagation -- hence the name of their method.
The hyperparameter $\eta>0$ is chosen using a validation set, which is also often used to select the best performing model among those that were produced during the course of running SGD (as opposed to using the model obtained when the optimization procedure finishes).



\section{Experiments}
\label{s:exp}

We empirically evaluated the training objectives $f_{\mathrm{lamb}}$ and $f_{\mathrm{quad}}$ of \cref{eq:obj-lamb} and \cref{eq:obj-quad}, respectively,
on MNIST and on various UCI data sets (\cite{Dua2019}). 
Both training objectives pb\_lambda ($f_{\mathrm{lamb}}$) and pb\_quad ($f_{\mathrm{quad}}$) were compared with the BBB objective (\cite{blundell2015weight}) and with vanilla SGD with momentum optimizer, see results in \cref{fig:mnist}. 

\begin{figure}[t]
\bc
\includegraphics[width=0.45\textwidth]{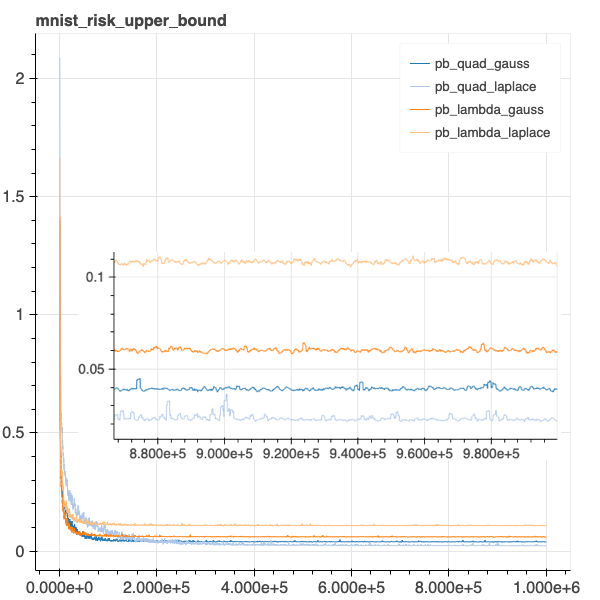}
\hfill
\includegraphics[width=0.45\textwidth]{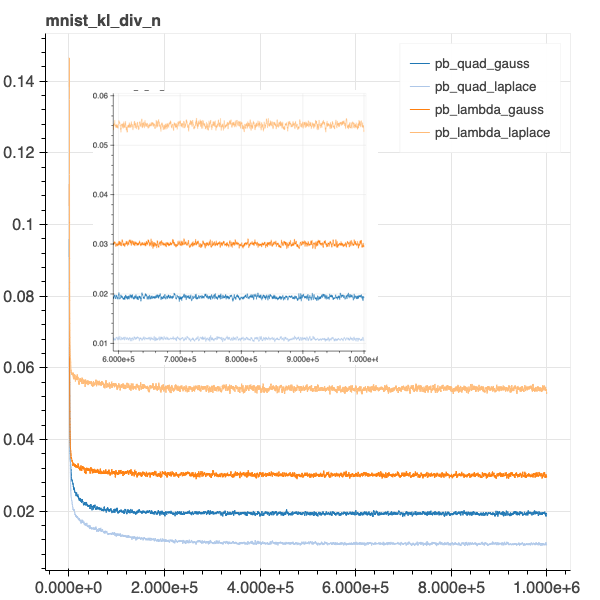}
\ec
\caption{Objectives `pb\_lambda' and `pb\_quad' with Gauss and Laplace distributions. Plots for risk upper bound and normalized KL divergence ($\KL(Q \Vert Q^{0})/ n_\text{train}$)  vs training iterations for MNIST. The KL divergence starts at $0.0$, increases quickly and then starts decreasing later in the training.}
\label{fig:mnist_prio} \vspace*{-3mm}
\end{figure}

We studied the effect of Gaussian and Laplace distributions over the model weights on MNIST for both our training objectives, using the same experimental setup as described in \cref{exp_MNIST}. 
Our prior distributions $Q^0$ were centered at the randomly initialized model weights.
The posterior distribution $Q$ is the same kind as the prior in each case. 
\cref{fig:mnist_prio} displays plots for the risk upper bounds and the normalized KL divergence ($\KL(Q \Vert Q^{0}) / n_\text{train}$). 
We note that `pb\_quad' with Laplace noise achieves minimum value of the risk upper bound and normalized KL. However, the test errors of `pb\_lambda' and `pb\_quad' with either distribution were similar.

In all the experiments on various data sets we performed a grid sweep over all the hyper parameters and then selected the run with the best risk upper bound. 
Model weights were initialized randomly from a truncated Gaussian distribution with standard deviation set to $1/\sqrt{n_\mathrm{in}}$, where $n_\mathrm{in}$ is the dimension of the inputs to a particular layer.
Our prior distributions $Q^0$ were centered at the randomly initialized weights.  
We did a grid sweep over the prior distribution scale parameter with standard deviation values in $[0.1, 0.09, 0.08, 0.07, 0.06, 0.05, 4\mathrm{e}-2, 3\mathrm{e}-2, 2\mathrm{e}-2, 1\mathrm{e}-2, 5\mathrm{e}-3, 1\mathrm{e}-3]$.
The variance of the posterior distribution was initialized to the same value as the prior distribution variance. We observed that higher variance leads to instability during training and lower variance does not explore the weight space. 
In all experiments we used SGD with momentum optimizer for training and performed a grid sweep over learning rate in $[1\mathrm{e}-3, 5\mathrm{e}-3, 1\mathrm{e}-2]$ and momentum in $[0.95, 0.99]$. We found that learning rates higher than $1\mathrm{e}-2$ caused divergence in training and learning rates lower than $5\mathrm{e}-3$ converged slowly.
We performed experiments using the cross entropy loss function during training, for which we enforced boundedness by restricting the minimum probability, and we did a grid sweep over $p_\mathrm{min}$ in the range $[1\mathrm{e}-2, 1\mathrm{e}-3, 1\mathrm{e}-4, 1\mathrm{e}-5, 1\mathrm{e}-8, 1\mathrm{e}-16]$. Values higher than $1\mathrm{e}-2$ distorts the input to loss function and leads to higher training loss. Whereas lower values did not seem to have any impact on training.
The lambda value in $f_{\mathrm{lamb}}$ was optimized using alternate minimization using SGD with fixed learning rate of $1\mathrm{e}-4$. 
Notice that BBB requires an additional sweep over a KL trade-off coefficient, 
which was done with values in $[1\mathrm{e}-6, 1\mathrm{e}-5, \ldots, 1\mathrm{e}-1]$, see \cite{blundell2015weight}. 
For BBB and our methods the predictions were obtained using the randomly sampled model weights.

\subsection{The choice of the prior distribution}
As pointed out above, our prior distributions (Laplace or Gauss) were centered at the randomly initialized weights. The attentive reader may object that, according to the usual PAC-Bayes theorems in the literature, priors are supposed to be non-random. To clarify, PAC-Bayes bounds require the prior to be a `data-free' (i.e. non data-dependent) distribution. The randomness in the initialization of the weights is external to the training process, hence it is of the `data-free' kind. This justifies that using priors centered at the randomly initialized weights is perfectly valid. 
Our choice of priors centred at the randomly initialized weights was motivated by observations of \cite{zhang2019all} indicating that post-training re-initialization of many weights has little effect on prediction performance. Subsequently we heard from \cite{dziugaite2017computing} that they used the same choice before.

\subsection{Experiments on MNIST}
\label{exp_MNIST}

\begin{figure}[t]
\bc
\includegraphics[width=0.45\textwidth]{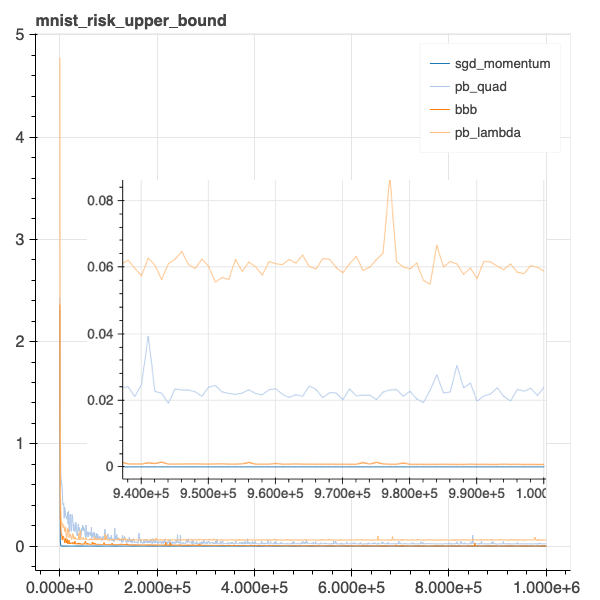}
\hfill
\includegraphics[width=0.45\textwidth]{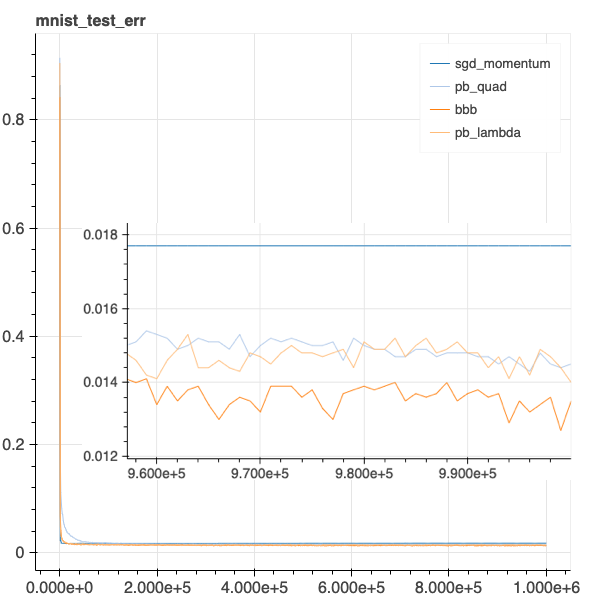}
\ec
\caption{Plots for training objectives and test errors vs the training iterations for MNIST.}
\label{fig:mnist}
\end{figure}

For experiments on MNIST we trained a feed forward neural network with 3 hidden layers, with 600 units and ReLU activations in each hidden layer. 
We trained our models using standard MNIST dataset split of 50000 training examples and 10000 test examples.
We ran all experiments for $1$ million training iterations. We observe from \cref{fig:mnist} that the methods converge around $100000$ iterations. We ran the experiments longer to check if the KL divergence can be minimized further. 
We used training batch size of $256$ for all the experiments. 

On \cref{fig:mnist} we report the test error
and the risk upper bound 
as a function on the number of training iterations. 
Our methods `pb\_lambda'  and  `pb\_quad'  achieve similar test accuracy at the end of training compared to BBB and SGD with momentum baseline. 
The $\eta$ parameter in  BBB which is of $O(1\mathrm{e}-5)$ makes the KL divergence component of the loss insignificant compared to train loss. Thus in case of BBB we see marginal decrease in the KL divergence during the course of training and the solution it returns is expected to be similar to that returned with the SGD with momentum baseline. The best test error achieved by our methods 0.014 is comparable to the risk upper bound of 0.02 at the end of training.
This is a significant improvement with respect to previously computed non-vacuous bounds.
Also we note that for `pb\_lambda' method the $\lambda$ value start at $1$ and decreases to $0.5$ and starts increasing again after around $150000$ iterations to finally reach a value of $0.75$.

\subsection{Comparison to  \cite{dziugaite2018data}}

We are not the first to propose training methods or risk certificates for neural networks based on PAC-Bayes bounds. \cite{LangfordCaruana2001} developed a method to train a stochastic neural network by randomizing the weights with Gaussian noise (adjusted via a sensitivity analysis) and computed an upper bound on the error using a PAC-Bayes bound. They run experiments on a UCI data set and a synthetic data set, and also pointed that the PAC-Bayes approach might be more fruitful for computing non-vacuous bounds.
\cite{dziugaite2017computing} derived a training objective from a similar PAC-Bayes bound to train a stochastic neural network with Gaussian randomization, and computed non-vacuous bounds on a `Binary MNIST' dataset. 
\cite{London2017} approached the generalization of stochastic neural networks by a stability-based PAC-Bayes analysis, and run experiments on CIFAR-10. Other efforts showing further evidence in favour of the PAC-Bayes approach for neural networks are e.g. \cite{neyshabur2017exploring,neyshabur2017approach}, and \cite{pitas2017}.

\begin{table}[htb]
\begin{center}
\renewcommand{\arraystretch}{1.2}
 \begin{tabular}{| c| c| c| c|} 
 \hline
Method & Test Error & Risk upper bound \\
\hline
\hline
\cite{dziugaite2018data} ($\tau=3\mathrm{e}+3$)  & 0.12 & 0.21 \\
 \hline 
\cite{dziugaite2018data} ($\tau=1\mathrm{e}+5$)  & 0.06 & 0.65 \\
 \hline
\cite{lever-etal2013} ($\tau=3\mathrm{e}+3$) &	0.12 & 0.26 \\ 
 \hline
\cite{lever-etal2013} ($\tau=1\mathrm{e}+5$) &	0.06 & 1 \\ 
 \hline
PBB & 0.014 & 0.023 \\
 \hline
\end{tabular}
\end{center}
\caption{Test set error / Risk upper bound for standard MNIST dataset.}
\label{tab:mnist}
\end{table}
\vspace{-2mm}



Usually, as we also noticed, the problem is that the KL term tends to dominate and most of the work in training is targeted at reducing it.
To address this issue distribution-dependent (\cite{lever-etal2013}) or data-dependent (\cite{dziugaite2018data}) priors can be used. 
 The latter derived a novel PAC-Bayes bound under differentially private (or, distributionally stable) priors
 and proposed a procedure to minimize the terms in the bound in connection to training neural networks. While there is no fully rigorous argument for backing up their finite-time procedure, an asymptotic argument is available.
What makes their work especially relevant to us is that they also provide results of experiments on standard multiclass MNIST. 
The values in \cref{tab:mnist} were reported by them.
We compare our results (PBB) with the best value they achieved on the test error and risk upper bound.

The parameter $\tau$ in their setting controls the temperature of a Gibbs distribution.  
In the table we display only the two values of their $\tau$ parameter which achieve best test error and risk upper bound.
Notice that our method did not use data-dependent priors, while \cite{dziugaite2018data} used data-dependent priors.
The rows `\cite{lever-etal2013}' display  values achieved by minimizing a PAC-Bayes bound that uses a different approach to relate the priors to the data, namely, the priors used by \cite{lever-etal2013} were distribution-dependent\footnote{The relevant part is their treatment of stochastic exponential weights prediction schemes, although they also treated stochastic RKHS methods with Gaussian randomization.} (see also \cite{lever-etal2010}). 
The method based on the PAC-Bayes bound of \cite{lever-etal2013} was implemented for neural networks by \cite{dziugaite2018data} and used as a baseline to contrast the experimental results.

It is interesting to compare our results with the values obtained by other methods in table 
above.
We note that \cite{dziugaite2018data}'s best values correspond to test accuracy of 94\% or 93\% while in those cases their bound values, although non-vacuous, were far from tight. On the other hand the tightest value of their bound only gives an 88\% accuracy.
By contrast, our method PBB achieves close to 98.6\% test accuracy (or 0.014 test error) even without using a data-dependent prior.
At the same time, as noted above, our optimal upper bound value (0.023) is much tighter than theirs (0.21).

\subsection{Comparison to  \cite{dziugaite2017computing}}

To compare our results with \cite{dziugaite2017computing}'s results we also run experiments on their `Binary MNIST' dataset where digits $0-4$ were mapped to class $0$ and digits $5-9$ were mapped to class $1$. We implemented the network architecture which achieved the best test error in their setting. 

\begin{table}[ht]
\begin{center}
 \begin{tabular}{||c| c| c|} 
 \hline
 Method & Test error & Risk upper bound  \\ [0.5ex] 
 \hline\hline

\cite{dziugaite2017computing} &  0.013 & 0.201 \\  \hline
PBB & 0.015 & 0.022 \\  \hline
\end{tabular}
\end{center}
\caption{Test set error / Risk upper bound for `Binary MNIST' dataset.} 
\label{tab:binary_mnist}
\end{table} 
\vspace{-2mm}

From the results in \cref{tab:binary_mnist} we note that 
we achieve similar test set accuracy, while the minimal value of our risk upper bound is significantly tighter, i.e. our method PBB further tightens the gap between the risk certificate (upper bound) and the risk estimate evaluated on a test set.

\section{Conclusion and Future Work}
\label{s:conclude}

We explored the `PAC-Bayes with Backprop' (PBB) methods to train probabilistic neural networks.
The take-home message is that these training methods are derived from sound theoretical foundations, and output models that come with a performance guarantee at no extra cost, since PBB training objectives are based on PAC-Bayes bounds which upper bound the risk. This is an improvement over methods derived heuristically rather than from theoretically justified arguments, and over methods that do not include a risk certificate that is valid on unseen examples.

We presented two PBB training objectives.
As far as we are aware, we are the first to use these objectives for neural network training.
The results of our experiments on MNIST have showed that these two PBB objectives give predictors with competitive test set performance and with computed non-vacuous bounds that are significantly improved compared to previous results. 
This indicates that PBB methods look promising for achieving self-bounding learning, since the values of the risk certificates output by the training methods are close to the values of the test set error estimates. Additional experiments on some UCI data sets confirm these observations.

However, since the results of initial experiments on CIFAR-10 showed risk certificates with values not as close to the test set error estimate as in the MNIST or UCI experiments, more needs to be done before `self-bounding learning' can really be claimed.
We found that the biggest issue when using PBB is the KL term. In particular this is noticeable for larger networks. Besides the obvious choice of smaller networks, some ideas to deal with this issue in future work are e.g. coupling of weights to reduce the number of terms in the KL, using hierarchical priors (mixture over networks of different sizes), or using data-dependent priors. We believe that these combinations may have the chance of giving tight bounds and simultaneously achieving state-of-the-art test set performance.

\bibliography{biblia}

\newpage
\appendix

\section{KL formulas: Laplace versus Gauss}
\label{s:L_vs_G}

The Laplace density with mean parameter $\mu \in \R$ and variance $b>0$ is the following: 
\begin{align*}
    p(x) = (2b)^{-1} \exp\bigl(-\frac{|x-\mu|}{b}\bigr)\,.
\end{align*}
The KL divergence for two Laplace distributions is as follows:
\begin{align}
    \KL(\operatorname{Lap}(\mu_1,b_1) \Vert \operatorname{Lap}(\mu_0,b_0))
    = \log(\frac{b_0}{b_1}) + \frac{|\mu_1-\mu_0|}{b_0} + \frac{b_1}{b_0}e^{-|\mu_1-\mu_0|/b_1} - 1\,.
\label{eq:KL_Laplace}
\end{align}

For comparison, recall that the Gaussian density with mean parameter $\mu \in \R$ and variance $b>0$ has the following form:
\begin{align*}
   p(x) = (2\pi b)^{-1/2} \exp\bigl(-\frac{(x-\mu)^2}{2b}\bigr)\,.
\end{align*}
The KL divergence for two Gaussian distributions is as follows:
\begin{align}
    \KL(\operatorname{Gauss}(\mu_1,\sigma_1^2) \Vert \operatorname{Gauss}(\mu_0,\sigma_0^2))
    = \frac{1}{2}\Bigl(\log(\frac{b_0}{b_1}) + \frac{(\mu_1-\mu_0)^2}{b_0} + \frac{b_1}{b_0}   - 1 \Bigr)\,.
\label{eq:KL_Gauss}
\end{align}

\section{Experiments on some UCI data sets}
\label{s:exp_UCI}

For experiments on UCI datasets we trained a feed forward neural network with ReLU activations in each hidden layer, with 64 units in the first hidden layer and 32 units in the second hidden layer, followed by an output layer. We performed experiments on 5 different UCI datasets. We ran all experiments for $30000$ training iterations. We used training batch size of $256$ for all the experiments.   

\begin{table}[ht]
\begin{center}
 \begin{tabular}{||c| c| c| c| c||} 
 \hline
 UCI dataset & pb\_lambda & pb\_quad & bbb & sgd\_momentum \\ [0.5ex] 
 \hline\hline

MUSHROOM &  0.0 / 0.0 & 0.0 / 0.0 & 0.0 & 0.0 \\  \hline
BREAST &  0.047 / 0.051 & 0.029 / 0.134 & 0.047 & 0.065 \\  \hline
AvsB &  0.0 / 0.013 & 0.0 / 0.052 & 0.0 & 0.0 \\  \hline
IONOSPHERE &  0.110 / 0.224 & 0.117 / 0.420 & 0.145 & 0.110 \\  \hline
SKIN &  0.001 / 0.001 & 0.001 / 0.001 & 0.002 & 0.002 \\  \hline
\end{tabular}
\end{center}
\caption{Test set error / Risk upper bound for UCI datasets.} \label{tab:uci_exp}
\end{table}
\vspace{-2mm}

In \cref{tab:uci_exp} we list the test error / risk upper bound  for various methods.
We note that both `pb\_lambda' and `pb\_quad' obtain similar test accuracy to that of BBB and SGD with momentum baselines. Also the risk upper bounds are comparable to the test errors. We would like to emphasize that BBB uses cross-validation 
over $\eta$ in \cref{eq:bbbobjective} to obtain good performance, while `pb\_lambda' automatically tunes the coefficient $\lambda$ which 
controls the weight of the various components in the objective function.

\end{document}

%% file: commands-omar.tex
\newcommand{\KL}{\operatorname{KL}}
\newcommand{\kl}{\operatorname{kl}}

\newcommand{\EE}{\operatorname{\mathbb{E}}} 

\newcommand{\R}{\mathbb{R}}  
\newcommand*{\quotientspace}[2]
{\ensuremath{\raisebox{.25ex}{\ensuremath{#1}} \hspace{-.35ex} \raisebox{-.25ex}{/} 
\hspace{-.05ex} \raisebox{-.85ex}{\ensuremath{#2}}} \hspace{.2ex} }
\newcommand*{\textquotientspace}[2]
{\ensuremath{\raisebox{.05ex}{\ensuremath{#1}} \hspace{-.30ex} \raisebox{-.15ex}{/} 
\hspace{.10ex} \raisebox{-.55ex}{\ensuremath{#2}}}}

\newcommand{\upsum}
{\ensuremath \mkern5.4mu \overline{\vphantom{S}\mkern8mu} \mkern-11mu S}
\newcommand{\lowsum}
{\ensuremath \mkern3.4mu \underline{\vphantom{S}\mkern8mu} \mkern-9mu S}
\newcommand{\upint}
{\ensuremath \mkern10.4mu \overline{\vphantom{\int}\mkern7mu} \mkern-21mu\int}
\newcommand{\upintwdomain}[1]
{\ensuremath \mkern10.4mu \overline{\vphantom{\int}\mkern7mu} \mkern-21mu\int_{#1}}
\newcommand{\textupint}
{\textstyle \mkern5mu \overline{\vphantom{\int}\mkern6mu} \mkern-14mu\intop\mkern0mu}
\newcommand{\textupintwdomain}[1]
{\textstyle \mkern5mu \overline{\vphantom{\int}\mkern6mu} \mkern-14mu\intop_{#1}}
\newcommand{\lowint}
{\ensuremath \underline{\vphantom{\int}\mkern7mu} \mkern-9mu\int}
\newcommand{\lowintwdomain}[1]
{\ensuremath \underline{\vphantom{\int}\mkern7mu} \mkern-9mu\int_{#1}}
\newcommand{\textlowint}
{\textstyle \mkern3mu \underline{\vphantom{\int}\mkern6mu} \mkern-8.5mu\intop}
\newcommand{\textlowintwdomain}[1]
{\textstyle \mkern3mu \underline{\vphantom{\int}\mkern6mu} \mkern-8.5mu\intop_{#1}}

\def\ddefloop#1{\ifx\ddefloop#1\else\ddef{#1}\expandafter\ddefloop\fi}

\def\ddef#1{\expandafter\def\csname b#1\endcsname{\ensuremath{\mathbf{#1}}}}
\ddefloop ABCDEFGHIJKLMNOPQRSTUVWXYZ\ddefloop

\def\ddef#1{\expandafter\def\csname bb#1\endcsname{\ensuremath{\mathbb{#1}}}}
\ddefloop ABCDEFGHIJKLMNOPQRSTUVWXYZ\ddefloop

\def\ddef#1{\expandafter\def\csname c#1\endcsname{\ensuremath{\mathcal{#1}}}}
\ddefloop ABCDEFGHIJKLMNOPQRSTUVWXYZ\ddefloop

\def\ddef#1{\expandafter\def\csname s#1\endcsname{\ensuremath{\mathsf{#1}}}}
\ddefloop ABCDEFGHIJKLMNOPQRSTUVWXYZ\ddefloop

\def\ddef#1{\expandafter\def\csname v#1\endcsname{\ensuremath{\boldsymbol{#1}}}}
\ddefloop ABCDEFGHIJKLMNOPQRSTUVWXYZabcdefghijklmnopqrstuvwxyz\ddefloop

\def\ddef#1{\expandafter\def\csname v#1\endcsname{\ensuremath{\boldsymbol{\csname #1\endcsname}}}}
\ddefloop {alpha}{beta}{gamma}{delta}{epsilon}{varepsilon}{zeta}{eta}{theta}{vartheta}{iota}{kappa}{lambda}{mu}{nu}{xi}{pi}{varpi}{rho}{varrho}{sigma}{varsigma}{tau}{upsilon}{phi}{varphi}{chi}{psi}{omega}{Gamma}{Delta}{Theta}{Lambda}{Xi}{Pi}{Sigma}{varSigma}{Upsilon}{Phi}{Psi}{Omega}\ddefloop

